\documentclass[letterpaper, 10pt, conference]{ieeeconf}    
\IEEEoverridecommandlockouts
\usepackage{cite}
\usepackage{amsmath,amssymb,amsfonts}
\usepackage{algorithmic}
\usepackage{graphicx}
\usepackage{textcomp}
\usepackage{xcolor}
\usepackage{cite}
\usepackage{gensymb}
\usepackage{tabularx}
\usepackage{multicol}
\def\BibTeX{{\rm B\kern-.05em{\sc i\kern-.025em b}\kern-.08em
    T\kern-.1667em\lower.7ex\hbox{E}\kern-.125emX}}
    
\makeatletter
\def\BState{\State\hskip-\ALG@thistlm}
\makeatother

\title{\bf Towards the Development of a Tendon-Actuated Galvanometer for Endoscopic Surgical Laser Scanning
} 

\author{Kent K. Yamamoto\textsuperscript{a,b}, Tanner J. Zachem\textsuperscript{a,b}, Behnam Moradkhani\textsuperscript{c}, Yash Chitalia\textsuperscript{c}, and Patrick J. Codd\textsuperscript{a,b}
\thanks{\textsuperscript{a}Department of Mechanical Engineering and Materials Science, Duke University, Durham, NC, USA} 
\thanks{\textsuperscript{b}Department of Neurosurgery, Duke University School of Medicine, Durham, NC, USA} 
\thanks{\textsuperscript{c}Speed School of Engineering, University of Louisville, Louisville, KY, USA}
\thanks{Corresponding author: Kent K. Yamamoto (\texttt{kky7@duke.edu})}
}

\begin{document}
\maketitle


\begin{abstract}

 There is a need for precision pathological sensing, imaging, and tissue manipulation in neurosurgical procedures, such as brain tumor resection. Precise tumor margin identification and resection can prevent further growth and protect critical structures. Surgical lasers with small laser diameters and steering capabilities can allow for new minimally invasive procedures by traversing through complex anatomy, then providing energy to sense, visualize, and affect tissue. In this paper, we present the design of a small-scale tendon-actuated galvanometer (TAG) that can serve as an end-effector tool for a steerable surgical laser. The galvanometer sensor design, fabrication, and kinematic modeling are presented and derived. It can accurately rotate up to 30.14 $\pm$ 0.90$\degree$ (or a laser reflection angle of 60.28$\degree$). A kinematic mapping of input tendon stroke to output galvanometer angle change and a forward-kinematics model relating the end of the continuum joint to the laser end-point are derived and validated.
\end{abstract}

\section{Introduction} 

\subsection{Clinical Background}
Neurosurgery is immensely delicate, as many neuroanatomical structures must be avoided. Due to the inability to precisely discern tumor margins and the need to navigate very tight working spaces, innovation is necessary to improve the surgeon's capability to operate minimally invasively utilizing endoscopic tools. Current applications of endoscopes in neurosurgery are most common for endonasal approaches to skull base lesions, minimally invasive craniotomies, and ventricular approaches. In all of these settings, the added view of the endoscope provides benefits, but there are two relevant drawbacks\textcolor{black}{\cite{endoscopyLim,endoscopyLim2}}. First, tissue information is limited to the color screen the surgeon is viewing. Second, the surgeon's ability to manipulate, cut, or cauterize said tissue is limited by linear tools, only allowing surgeons to access the space axial to the tool.

Within neurosurgery, fluorescence-guided surgery is a popular option for increased tissue identification, as specific pathologies have been shown to improve outcomes \cite{Mahboob}. However, exogenous fluorophores, delivered to the patient as bio-molecules/pro-drugs are required and only target specific tumor grades and types \cite{5ALA,NF}. Our group has previously validated an endogenous laser-induced fluorescence spectroscopy device, the TumorID, that is applicable across tumor types\cite{TIDTucker}. Specifically, it has shown the ability to discern between pituitary adenoma subtypes and tumor tissues from healthy tissue\cite{TZID}. Regarding tissue manipulation, higher energy lasers can coagulate and ablate tissue, which could assist neurosurgeons in tight approaches when deploying a bulky, mechanical tool is difficult\cite{NSGYlaser}. 

\begin{figure}[h!]
    \centering\includegraphics[width=\linewidth,keepaspectratio]    {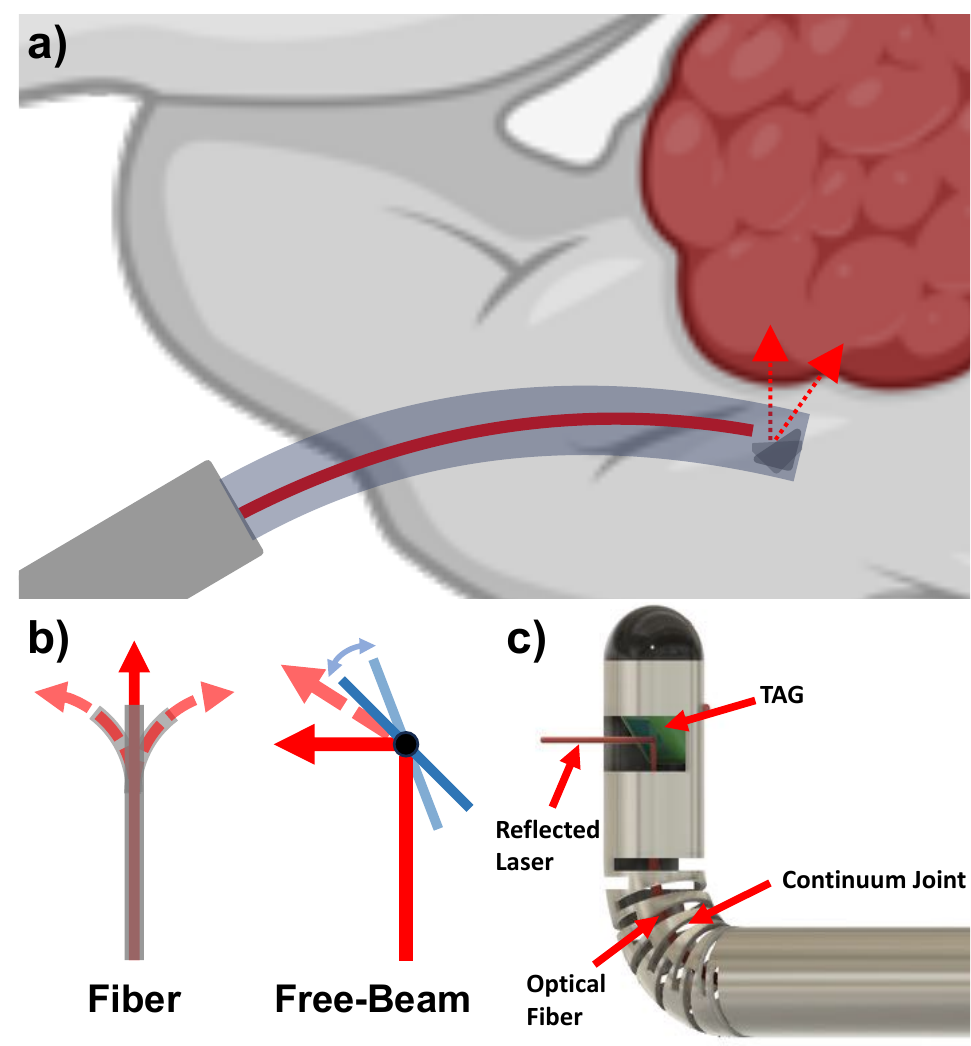}
    \caption{a) Rendition of proposed TAG approaching a pituitary lesion to aid in resection. Note the tool's ability to both fiber and free-beam steer to reach otherwise limited locations on the lesion \cite{Biorender}. b) Visual representations of the fiber and free-beam steering modalities. c) Rendition of proposed steerable surgical laser tool with fiber and free-beam steering capabilities.}
    \label{fig:intro}
\end{figure}

The use of lasers and robotics in neurosurgery is growing. Laser interstitial thermal therapy (LITT) comprises a neuronavigation robot and a small laser fiber to enter the core of a tumor for epileptic focus with MRI thermometry. Heating the lesion from the inside out induces cell death and dysfunction \cite{LITT}. LITT can be a benchmark for laser and robotic-based systems that can be adopted into the neurosurgical workflow. There are potentially new applications for lasers in surgery, especially for the resection of tumors from the endonasal approach, as illustrated in Fig.~\ref{fig:intro}-a.

\subsection{Laser Steering Modalities}

In surgical applications, laser steering can be categorized into fiber and free-beam steering \cite{laserReview}, illustrated in Fig.~\ref{fig:intro}-b. The former refers to a system in which  light travels along an optical fiber from a source to the distal tip of the fiber. By bending the fiber within its maximum bend radius range, the flexibility of the fiber allows light to traverse through irregular geometry within a small form factor \cite{bendRadius}. \textcolor{black}{By inserting an optical fiber through the lumen of continuum robots such as tendon-actuated notched joints or concentric tube systems, the optical fiber can be steered in a controlled manner.} Recent work in surgical robotics has utilized the flexibility in optical fibers to deliver light or energy at the end of endoscopic robots for laser cutting and ablation \cite{hybridLaser, ficheraLarynx, hsmrLITT}. Endoscopic imaging that requires optical fibers at the end-effector (illumination for chip-on-tip cameras, OCT, etc.) also utilizes fiber-steering for image acquisition \cite{ficheraCHIP,bernardOCT}. The steerability of continuum and endoscopic robots pairs well with optical fibers, thus allowing for transmitting light (and, in general, energy) through tortuous anatomical structures. The workspace is also increased due to the precise control of the end-effector when an optical fiber is sheathed within a steerable continuum joint. However, some drawbacks with fiber steering include lower laser-point steering speed and moving the entire robotic joint to steer the laser. 

Free-beam steering is controlled by both static mirrors and galvanometers - rotating mirrors. Coupling two galvanometers rotating about different axes allows laser point steering in a plane. Laser steering with galvanometers provides faster and greater reachability within the scanning plane. Autonomous tissue resection via free-beam laser has been successful in a bench-top setting \cite{tumorCNC,stereoCNC}, and free-beam surgical laser tools, such as laser scalpels, are already in clinical use. Drawbacks with free-beam steering include linear trajectories and preventing laser steering along a curved path. The optical components that come with free-beam steering, such as galvanometers, also require more space than a compact end-effector package, which is possible with fiber steering.

To combine the advantages of fiber and free-beam steering, York et al. have developed a printed-circuit micro-electromechanical (PC-MEMS) galvanometer that attaches to the end of a colonoscope \cite{york}. This allows for complex navigation using the colonoscope and precise, rapid laser point steering while the scope is fixed. Although the proposed galvanometer attachment allows for 18~mm x 18~mm rapid planar scanning, the attachment adds length to the distal end of the scope, and the linkage system for each rotating mirror has many components, increasing mechanical complexity.
\subsection{Proposed Solution}

Inspired by \cite{york}, we propose a novel micro-galvanometer system that is tendon-actuated, thus able to be integrated within a continuum joint, as shown in Fig. \ref{fig:intro}-c. The actuating wire and the optical fiber will be routed through the lumen of the continuum joint. A rotating mirror requires only one wire, simplifying the mechanical design and streamlining the actuation method. By attaching a galvanometer at the tip of the continuum joint, the steerable laser tool can both fiber and free-beam steer. The tool can navigate to the target region, then actuate the galvanometer to accurately and precisely scan the region with the laser without moving the tool.

The structure of this paper is as follows: We first propose the design of the tendon-actuated galvanometer (TAG) and present a geometric model relating mirror angle to tendon stroke and a forward-kinematics model of the end laser point in Section \ref{sec:MM}. In Section \ref{sec:results}, we present results from a bench-top experiment to evaluate the geometric and forward-kinematics models. Finally, we conclude the paper with further reasoning of the results, possible sources of error, new design considerations, and future directions with the TAG. 

\section{Materials and Methods} \label{sec:MM}
\subsection{Tendon-Actuated Galvanometer (TAG)} 

\begin{figure}[h!]
    \centering\includegraphics[width=\linewidth,keepaspectratio]{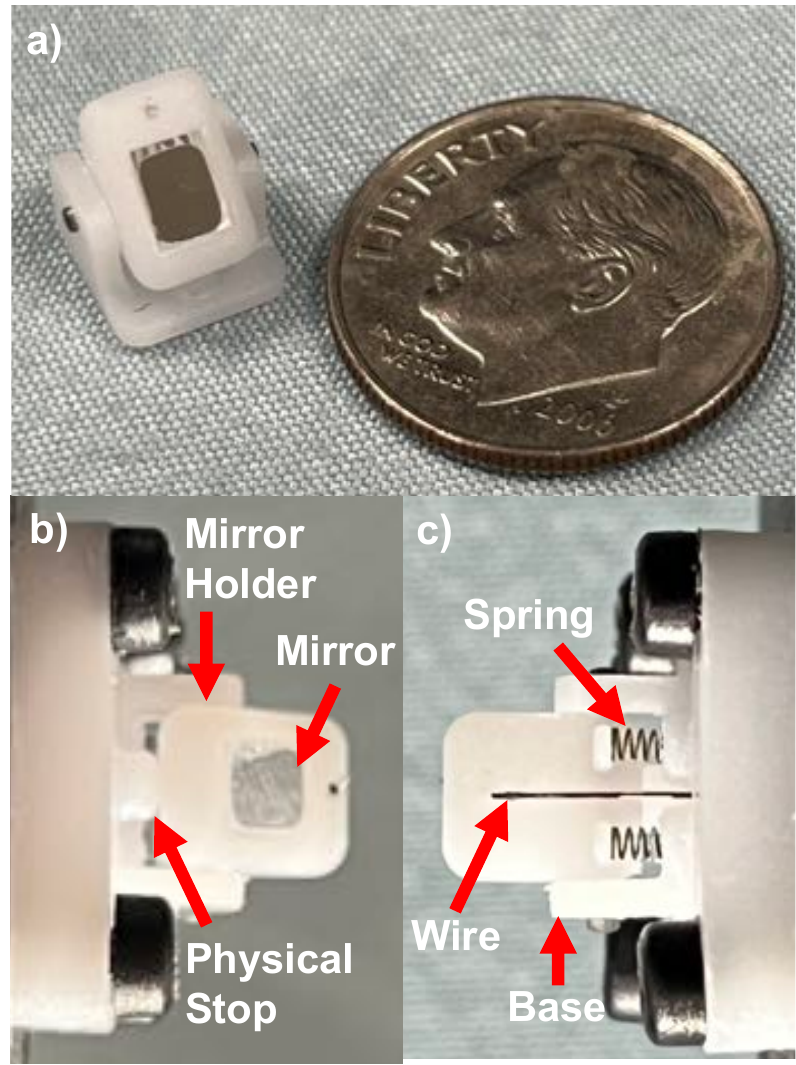}
    \caption{a) Assembled TAG in comparison to a U.S. dime (approximately 18~mm diameter). b) Front view of the TAG. c) Back view of the TAG.}
    \label{fig:TAG}
\end{figure}

The assembled TAG, shown in Fig. \ref{fig:TAG}-a, is an 8~mm x 8~mm x 7~mm end-effector tool for steerable surgical tools. It consists of 5 key components: the mirror, mirror holder, base, two springs, and a wire, shown in Fig. \ref{fig:TAG}-b and \ref{fig:TAG}-c. The mirror holder and base are designed in CAD software and 3D-printed (Projet 2500 Plus MJP, 3DSystems, Rock Hill, SC, USA). The geometry of the mirror holder is based on the mirror implemented - a Right-Angle Prism Mirror with a Broadband Dielectric Coating ($R_{avg}=99\%$) for 400-750~nm (MRA03-E02, ThorLabs, Newton, NJ, USA). This mirror is selected for its small size, low cost, broadband optic coating, and availability with laser line coatings. Optical adhesive (Norland Optical Adhesive NOA 68, Jamesburg, NJ, USA) rigidly secures the mirror to the mirror holder. The compression springs (CB0040B 01 E, Lee Springs, Brooklyn, NYC, USA) are fixed in their respective slots on the base. An 8 mm long steel dowel (McMaster-Carr, Elmhurst, IL, USA) is inserted through the base and the mirror holder, securing them together and allowing the mirror holder to rotate about the dowel. Finally, a $0.007"$ outer-diameter (OD) Nitinol (nickel-titanium alloy) wire is fed through the holes in both the mirror holder and base. Each wire's end is attached to the galvanometer and actuation system. 

The TAG is a single-wire system that translates linear actuation to rotational motion. The actuation scheme is inspired by a tendon-actuated surgical grasper developed in \cite{tjGrasper}, where reversing the motion of actuation is achieved by implementing a compression spring. By slacking the wire, the spring force will allow the galvanometer to restore to its original configuration defined by the physical stop. The spring mentioned above is chosen for its low solid height (the height of the spring at max compression) of 0.81~mm to allow maximum compression, resulting in a greater change in the galvanometer angle. The current design consists of two springs to implement equal spring force on both sides of the pulled wire, and the proposed model does not account for the spring dynamics.

\subsection{TAG Modeling}

The TAG kinematics model input is tendon stroke, $t_{s}$, and the output is the laser incident angle, $\delta$. Thus, the initial configuration of the TAG ($t_{s} = 0$) will result in an incident angle of $45$$\degree$ due to the geometry of the prism mirror:
\begin{equation}
    \delta(t_s) = 45\degree + \phi(t_s)
    \label{incident}
\end{equation}
where $\phi$($t_{s})$ is the angle between the physical stop of the mirror holder, shown in Fig. \ref{fig:methods}. The base of the mirror holder can be modeled as a simple lever, illustrated in Fig. \ref{fig:methods}-b, and the following equation for $t_s$ can be formed:

\begin{figure}[h!]
    \centering\includegraphics[width=\linewidth,keepaspectratio]{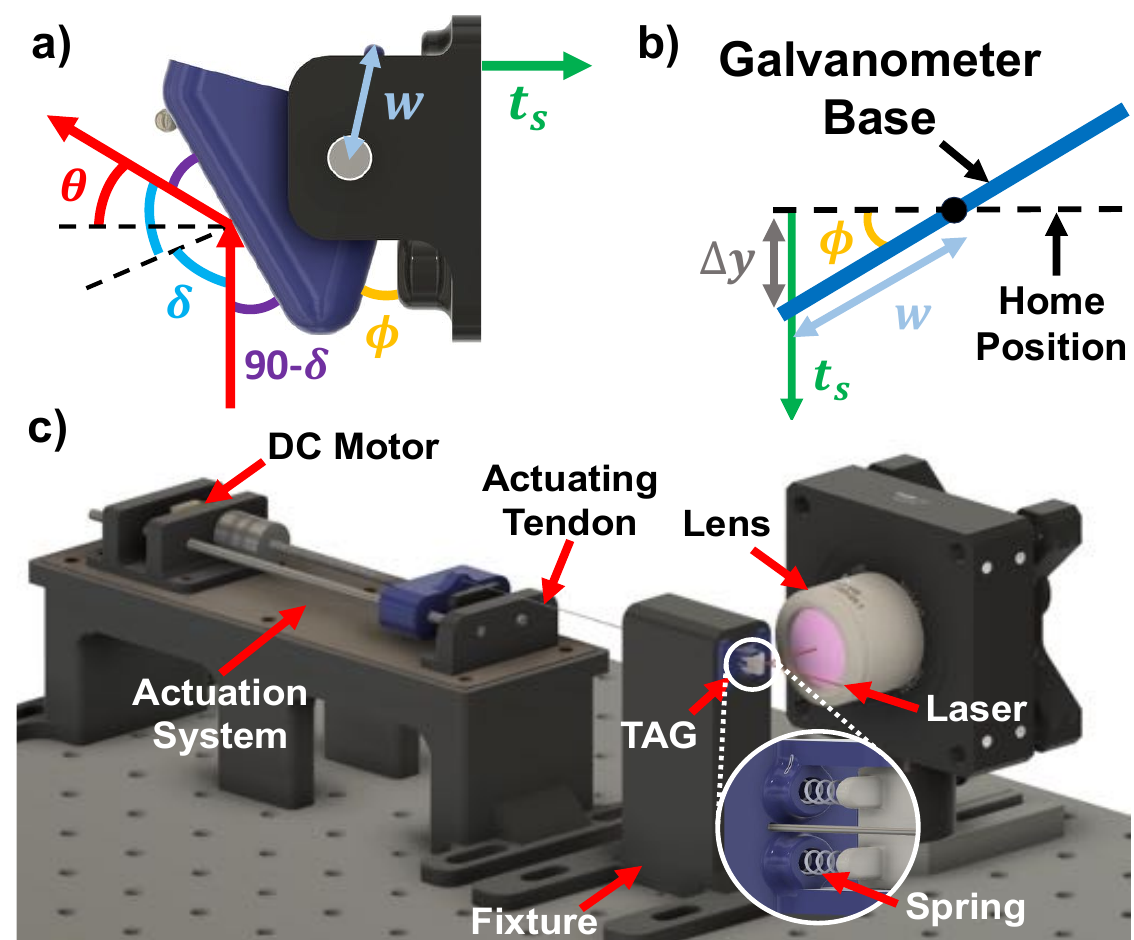}
    \caption{a) Schematic of TAG with variables relevant to the geometric and forward-kinematics modeling. b) Geometric modeling of TAG to obtain relationship between $\phi$ and $\Delta y$. c) Rendering of the experimental setup illustrating the actuation system, TAG, and laser assembly.}
    \label{fig:methods}
\end{figure}

\begin{equation}
    t_s = \Delta y + e
    \label{baseStroke}
\end{equation}
where $\Delta y$ is the change in length due to spring compression and $e$ is a tendon elongation factor. A relationship between spring compression length and galvanometer angle can be determined through geometric mapping:

\begin{equation}
    \Delta y = l\sin\left(\phi(t_s)\right)
    \label{geometric}
\end{equation}
where $l$ is the fulcrum length ($w$ = 2.83~mm). $e$ can then be determined by rearranging the Young's Modulus equation \textcolor{black}{(relating stress and strain)} and deriving the tendon elongation component for a wire with a circular cross-section under the force of the two Galvanometer springs in parallel:

\begin{equation}
    e = \frac{\sigma L}{E} = \frac{2K_s t_s L}{E\pi r^2}
    \label{elongation}
\end{equation}
where $K_s$ is the spring constant ($K_s$ = 0.269~$\text{N}/\text{mm}$, $w$ is the original length of the wire ($w$ = 142~mm), $E$ is the Young's Modulus of the wire ($E$ = 53.97~GPa from \cite{tendonE}), and $r$ is the radius of the wire ($r$ = 0.178 ~mm). Combining Eqs. (\ref{baseStroke}), (\ref{geometric}), and (\ref{elongation}) result in:

\begin{equation}
    t_s = l\sin(\phi(t_s)) + \frac{2K_s t_s L}{E\pi r^2}
    \label{result}
\end{equation}
We can then rearrange Eq. (\ref{result}) to solve for $\phi (t_s)$ and plug into Eq. (\ref{incident}) for the final kinematic model that relates tendon stroke to incident angle:

\begin{equation}
    \delta(t_s) = 45\degree + \arcsin\left(\left(1 - \frac{2K_s L}{E\pi r^2}\right)\frac{t_s}{l} \right)
\end{equation}

\begin{figure}[h!]
    \centering\includegraphics[width=\linewidth,keepaspectratio]{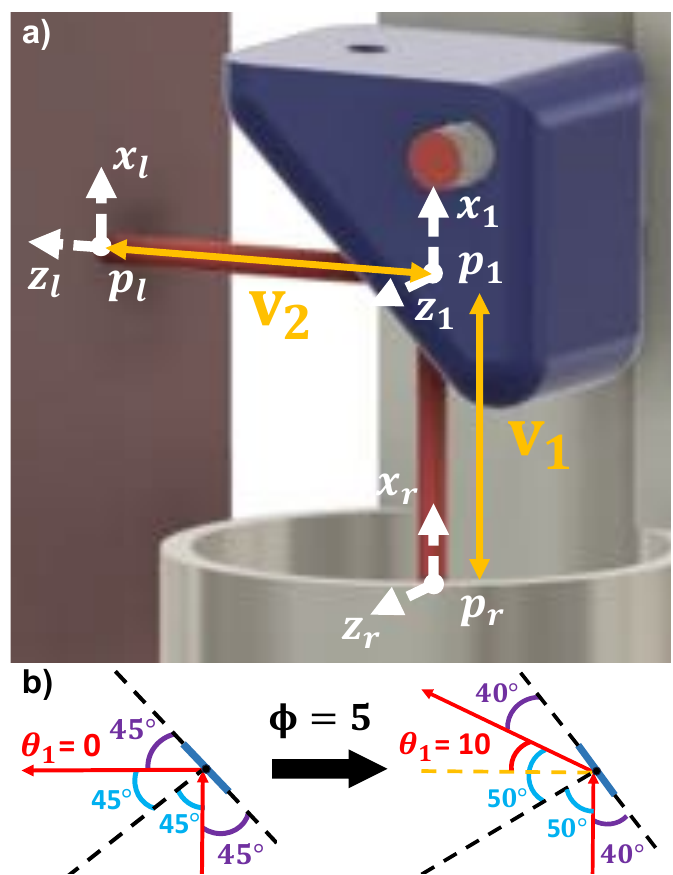}
    \caption{a) Defined coordinate system used for deriving homogeneous transformation matrix to obtain end laser point, $p_{l}$, with respect to the end of the continuum joint, $p_{r}$. b) Diagram illustrating that the laser angle with respect to the global horizontal will equal twice the TAG rotation angle.}
    \label{fig:dh}
\end{figure}

\subsection{TAG Forward Kinematics}

Assuming a known homogeneous transformation matrix, $H^{0}_{r}$, that relates the robot base to the end of the continuum joint, a homogeneous transformation matrix from the end of the continuum joint, $p_{r}$, to the end-laser point, $p_{l}$, can be derived. By treating the laser segments v$_1$ and v$_2$ depicted in Fig. \ref{fig:dh}-a as links, Denavit-Hartenberg (DH) parameters can be constructed as follows:

\begin{table}[!h]
\begin{center}
\caption{\label{DH}DH Parameters for TAG Forward Kinematics}
\begin{tabular}{| c || c | c | c | c |} 
 \hline
 $\#$ & $\theta$ & $d$ & $a$ & $\alpha$ \\ 
 \hline\hline
 $r$ & - & - & v$_1$ & 0 \\ 
 \hline
 $1$ & $\theta_1$ & 0 & 0 & $-\cfrac{\pi}{2}$ \\
 \hline
 $l$ & $0$ & $\cfrac{\mbox{v$_2$}}{c_{\theta_1}}$ & $-$ & $-$ \\
 \hline
\end{tabular}
\end{center}
\end{table}
Note that the $d$ value for the $l$ row in Table \ref{DH} has a $c_\theta$ term in the denominator. v$_2$ reaching the scanning surface requires this correction, as the laser propagates axially to the wall surface (compared to a definite-length rigid link). Therefore, the length of the scanning laser beam is appropriately coupled to the angle rotation of the galvanometer. The following transformation matrix relating $p_l$ to $p_r$, or $H^{r}_{l}(\theta_{1})$, can then be formed:

\begin{center}
$H^{r}_{l}(\theta_{1})$ = 
$\begin{bmatrix}
c_{\theta_{1}} & 0 & -s_{\theta_{1}} & v_{1} - v_{2}\tan{\theta_1} \\
s_{\theta_1} & 0 & c_{\theta_1} & v_2\\
0 & -1 & 0 & 0\\
0 & 0 & 0 & 1
\end{bmatrix}$
\end{center}
Due to the Law of Reflection, illustrated in \textcolor{black}{Fig. \ref{fig:dh}-b}, the input angle into the homogeneous transformation matrix, $\theta_1$, is twice the angle of the galvanometer input:

\begin{equation}
    \theta_{1} = 2(\phi(t_s))
\end{equation}
The top-right 3x1 vector of the homogeneous transformation matrix will then output the laser point position:
\begin{center}
$p^r_{l}$ =
$\begin{bmatrix}
v_{1} - {v_2}\tan{\theta_1}\\
v_2\\
0\\
\end{bmatrix}$
\end{center}
\subsection{Actuation System}

The TAG is actuated by a single nitinol wire at the mirror holder. The wire is routed through the TAG base and fixture, then attached to a 3D-printed actuation system shown in Fig. \ref{fig:methods}-c. The actuation system consists of a DC motor (Pololu Robotics and Electronics, Las Vegas, NV, USA) with a magnetic quadrature encoder that rotates a 0.6~mm (0.024") pitch lead screw with a linear rail. This will allow the wire to be translated by the wire stroke, $t_s$, mentioned in the kinematic modeling. 

\subsection{Laser Assembly}
The laser used to validate galvanometer functionality is a 635~nm 0.90~mW diode powered via USB (PL202, ThorLabs, Newton, NJ, USA). The laser is focused with a scan lens (EFL = 39~mm and WD = 25~mm, LSM03-VIS, ThorLabs, Newton, NJ), that produces a $1/e^2$ spot diameter of approximately 12.5~$\micro$m. The described laser assembly is also depicted in Fig \ref{fig:methods}-c.

\section{EXPERIMENTS AND RESULTS} \label{sec:results}

\subsection{TAG Actuation Experiment}

The experiment conducted is as follows: an image of the TAG, as depicted in Fig. \ref{fig:img}-a, is captured every 0.05 mm of wire displacement, up to 2~mm, for a total of 40 images. Each image is then processed to calculate the TAG angle by detecting the TAG edge and relating it to a global vertical vector. The experiment is run five times, and the average of the results is plotted with the theoretical model, including and excluding the wire elongation factor.

\subsection{Image Processing}
\begin{figure}[h!]
    \centering\includegraphics[width=\linewidth,keepaspectratio]{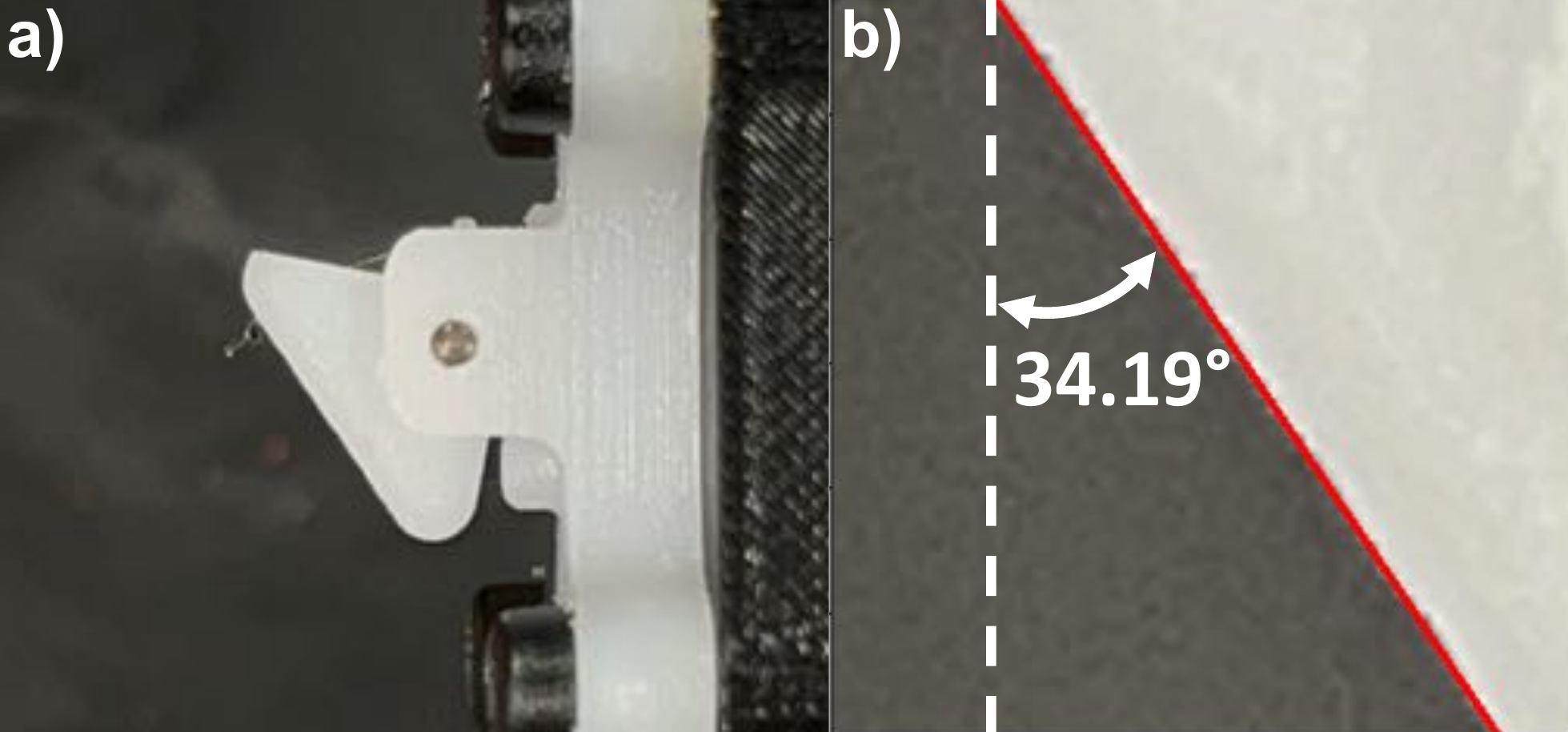}
    \caption{a) Image of TAG taken when tendon stroke is 0.5 mm. b) Cropped image of TAG with detected edge and calculated angle.}
    \label{fig:img}
\end{figure}

To automatically determine the mirror's angle, image processing is applied to find the edge. The process comprises four major steps: cropping, thresholding, edge filtering, and fitting a line to the points on the edge. The angle is calculated and compared to the model's prediction from this line of best fit. Cropping is conducted so that the only portion of the image processed is the mirror holder itself. This allows for efficient edge detection, as the only significant edge in the scene is the one of interest. Next, since the scene background is black and the mirror housing is white, a basic thresholding step is conducted. The image is converted to grey scale, then any pixel intensity, $I_{r,c}$, is mapped to a new pixel, $J_{r,c}$, by $I_{r,c} > 125 \rightarrow J_{r,c} = 255 $, $I_{r,c} \leq 125 \rightarrow J_{r,c} = 0$. At this point, a Canny filter is used to find pixels on the edge, and parameters of threshold$_1$ = 100, threshold$_2$ = 150, and aperture size = 7 were selected after tuning. Finally, to determine the angle of the mirror, linear regression is conducted on the points identified as being on the edge. This allows for a robust description of the edge. The angle of the mirror can be calculated using:
\begin{equation}
    \delta=90^{\degree}-\arctan\left(\left|\frac{1}{m}\right|\right)
\end{equation}
where $m$ is the slope of the linear regression line. Fig. \ref{fig:img}-b shows the resulting processed image and calculated angle. 

\subsection{Geometric Model Validation}
\begin{figure}[h!]
    \centering\includegraphics[width=\linewidth,keepaspectratio]{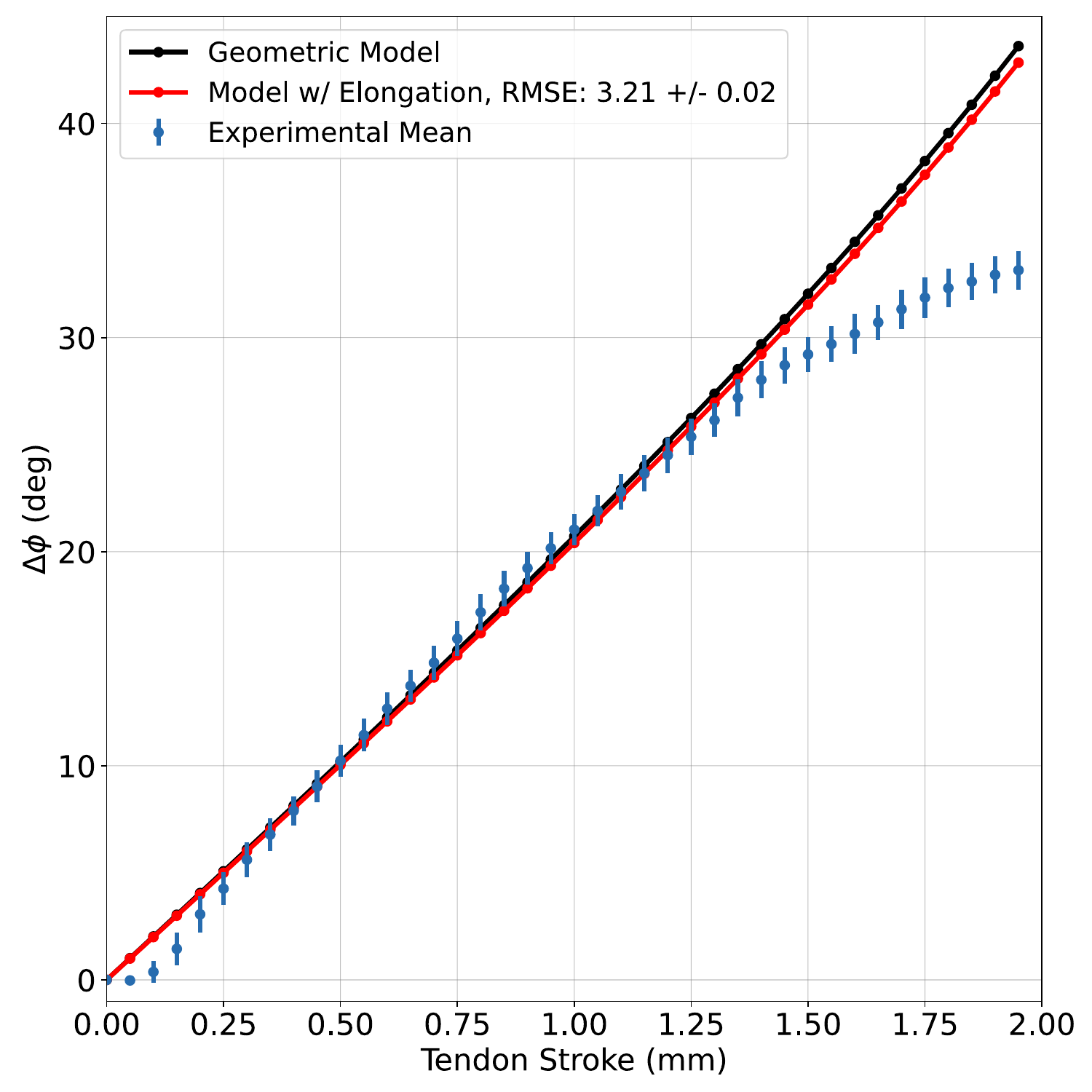}
    \caption{Tendon stroke plotted against average change in angle plotted for the geometric model, model with elongation, and the experimental result.}
    \label{fig:results}
\end{figure}

Due to the nature of 3D printing and tolerancing, the starting position of the TAG is not precisely horizontal. To account for this, the change in angle due to tendon stroke, $\Delta\theta$, is calculated and used to validate the geometric model. This is done by subtracting the starting angle from the other 39 calculated angles for both the theoretical and physical models. The average results of the five trials are shown in Fig. \ref{fig:results}, reporting an RMSE of 3.51~$\pm$ 0.01$\degree$. The graph shows a slight deviation from the model at the beginning but reasonably follows the model until a tendon stroke of approximately 1.25~mm. The experimental data then noticeably deviates from the model, reporting $\Delta\theta$ values slightly lower than the model. The difference between the model and experimental data also increases at greater tendon strokes. Finally, there seems to be no significant difference when comparing the theoretical models with and without the elongation factor.

\subsection{TAG Laser Steering}
The laser assembly is aimed at the galvanometer, and the laser is turned on. The beam is then reflected onto a white surface perpendicular to the actuation system, as shown in Fig. \ref{fig:fk}-a. The white surface is 8.56 mm away from the TAG. Using the validated geometric model, the TAG is rotated by 10, 20, and 30 degrees with the laser on. The change in laser end-point position, $\Delta x_{l}$ is measured at each angle with respect to the initial point position, as depicted in Fig. \ref{fig:fk}-b. Each experiment is run five times, and the average $\Delta x_{l}$ is calculated and compared to the translational x-value from the homogeneous transformation matrix for each angle input.

\subsection{TAG Forward Kinematics Validation}

\begin{table}[!h]
\begin{center}
\caption{TAG Laser Steering Results}
\begin{tabular}{| c | c | c | c |} 
 \cline{2-4}
 \multicolumn{1}{c|}{}&
 \multicolumn{3}{|c|}{$\theta_{1}$ ($\degree$)}\\
 \cline{2-4}
  \multicolumn{1}{c|}{}& $10$ & $20$ & $30$\\ 
 \cline{2-4}
 \hline
 Theoretical $\Delta x_{l}$ (mm) & 3.12  & 7.18  & 14.83 \\ 
 \hline
 Measured $\Delta x_{l}$ (mm)&  3.14$\pm$0.31 & 7.97$\pm$0.27 & 13.96$\pm$0.33 \\
 \hline
 $\%$ Error & +0.76 & +10.94 & -5.87 \\
 \hline
\end{tabular}
\label{table:1}
\end{center}
\end{table}

\begin{figure}[h!]
    \centering\includegraphics[width=\linewidth,keepaspectratio]{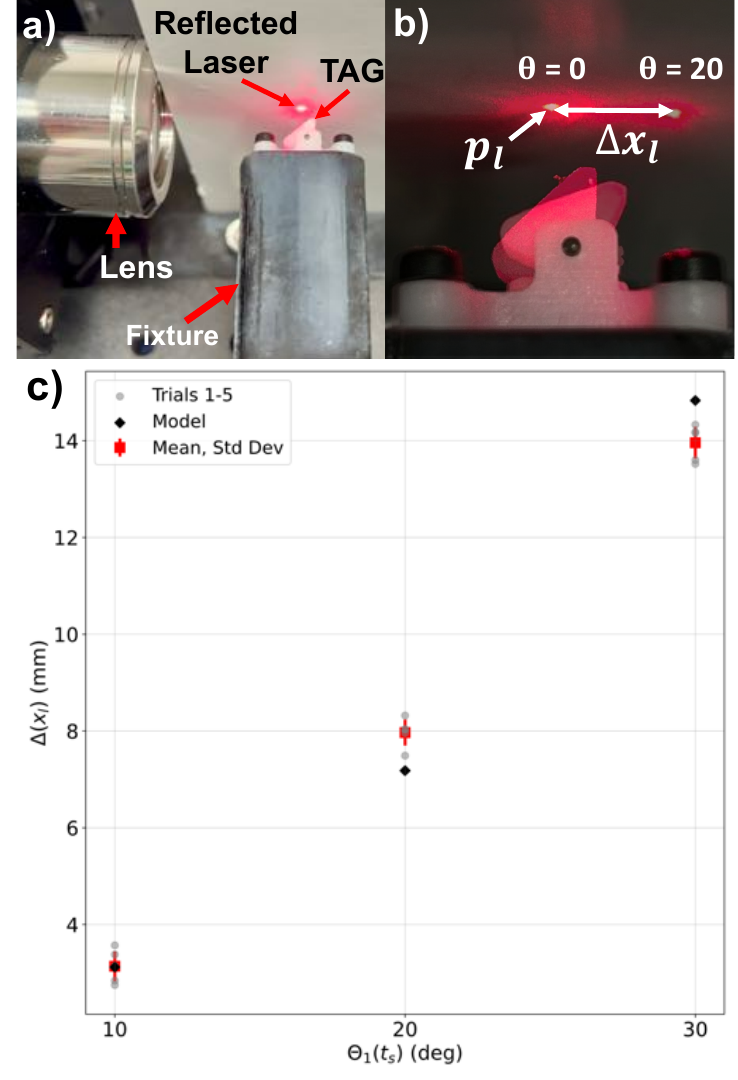}
    \caption{a) Experimental setup for laser steering study. b) Superimposed images of laser point, both at initial $\Delta\theta$ of 0 and 20 with laser on. c) Results from forward kinematics validation. $\Delta x$ for each trial and angle reported and compared against mean and model results.}
    \label{fig:fk}
\end{figure}

Table \ref{table:1} presents the theoretical $\Delta x_{l}$ value, the average and standard deviation of the measured $\Delta x_{l}$ value, and the error between the two for each of the three input angles. Fig. \ref{fig:fk}-c reports the $\Delta x_{l}$ values for each trial, the average of the five trials, and the model output per input angle.

The results from Table \ref{table:1} show that the average distance measured for the 10$\degree$ sweep is very close to the theoretical distance (\% error = $0.76$. The measured 20$\degree$ sweep distance overshoots the theoretical value, and the 30$\degree$ undershoots. These results align with the results from Fig. \ref{fig:results}, as the model and measured $\Delta\theta$ values at 10, 20, and 30 were similar, greater, and lower, respectively. The RMSE for the forward kinematics model is calculated to be 0.68~mm, implying that the model and experimental results are similar.

\section{DISCUSSION}

The experimental results from actuating the galvanometer showed similar outputs from the geometric model up to a tendon stroke of approximately 1.25 mm (or $\Delta\theta$ = 25$\degree$). In the first few millimeters of tendon stroke, the mirror does not rotate, which is hypothesized to be due to static friction. However, the mirror angle corrects towards 0.25 mm of stroke. After consistently following the predicted angles, the TAG does not rotate as much compared to the model after 1.25 mm of stroke. This could be due to elongation within the nitinol wire at greater forces, reducing the change in angle. Additionally, friction within the joints or between the surfaces of the 3D-printed parts may also result in less $\Delta\theta$ for higher stroke lengths. Within the two models plotted, there is a negligible difference when incorporating the wire elongation factor. However, increasing the spring stiffness will impact the strain of the wire, affecting how much elongation occurs. \textcolor{black}{Possible solutions to the deviation from model include reducing friction within the pivot point of the TAG and reduce the spring's stiffness, resulting in less overall force required to actuate the tendon.}

The forward kinematics validation experiment showed promising results, aligning with the results from Fig. \ref{fig:results}. The standard deviation of the measured scan length for each $\theta_{1}$ value was less than 0.35 mm, but this can be further reduced for more accurate laser steering. The error can be due to the lack of an exact 0\degree starting point, as tolerances are introduced from the 3D-printed physical stop for the mirror. To add, when the laser starts losing focus at an extreme angle \textcolor{black}{($\Delta\theta$ $>$ 30$\degree$)}, it is difficult to locate where the center of the laser point is; the circular point is now projected as an ellipse on the scanned surface.

\subsection{Limitation}

Although the current TAG design may fit in larger flexible surgical tools such as a colonoscope, it is too large to attach inside a notched continuum joint for neurosurgical applications (approximately 2 mm OD). This is due to the commercially available mirror used, which is 4 mm without the mirror holder, already too large for endoscopic neurosurgery. Additionally, the mirror can only rotate in one direction due to its wedged shape. With a flat planar mirror, bidirectional steering of a single TAG can be explored.

This study has some limitations when it comes to future optical implementation. Mainly, since the laser is scanning off of a mirror that rotates, future work implementing spectroscopy will need to account for the aberrations that will occur. The current design relies on a dichroic mirror, which will be swapped for a silver-coated mirror of the same size for spectroscopy applications to mitigate angle of incidence (AOI) specificity of dichroic coatings. Furthermore, the small form factor of this system and mirror will limit the maximum laser power it can deliver. However, through the use of laser line mirrors, cooling solutions, or single-use components, this can be overcome in the final design.

\subsection{Future Work}
The TAG introduces new trajectories with surgical laser research. First, the design of the galvanometer will be further optimized in size and accuracy. The current TAG design revolves around the smallest and most cost-efficient mirror available by the manufacturer. By customizing or fabricating smaller mirrors, the TAG can reduce in size to fit inside a neuroendoscopic continuum joint. This will enable the exploration of bidirectional, multi-modal surgical laser steering. Coupling two TAGs will also be explored, as it will allow 2D free-beam scanning with only two wires for actuation. Furthermore, our team believes a cutting laser (high-powered Nd:YAG fiber and laser line mirror) and optical coherence tomography (OCT) can be deployed and steered by TAG. We hope to explore this for minimally invasive precision laser imaging and cutting. Finally, our group will explore implementing a modified version of our TumorID with the TAG for minimally invasive tumor identification scanning.

\section{Conclusion}

This paper proposes a prototype of a novel miniature galvanometer that is tendon-actuated. A kinematics model utilizes tendon stroke to calculate galvanometer angle and is tested and validated in benchtop experiments, with an RMSE of 3.51 $\pm$ 0.01$\degree$. A forward-kinematics model of the end laser point is also derived and validated in a laser steering experiment, resulting in an RMSE of 0.68 $\pm$ 0.33 ~mm. The TAG will be further optimized in form factor and accuracy as an end-effector attachment for \textcolor{black}{continuum surgical robots for multimodal laser steering.} This will open new possibilities for lasers in surgery by introducing multi-modal laser steering in a single tool for increased maneuverability and precision.

\section{Acknowledgment}

The research reported in this publication was supported by the NSF-NRT Traineeship in Advancing Surgical Technologies (TAST). The authors would like to thank Dr. Brian Mann, Evan Kusa, members of the Brain Tool Lab and The HeART Lab, and TAST faculty and staff for their continued support and valuable feedback.

\bibliographystyle{IEEEtran}
\bibliography{main}

 \newcommand{\noop}[1]{}
\begin{thebibliography}{10}
\providecommand{\url}[1]{#1}
\csname url@samestyle\endcsname
\providecommand{\newblock}{\relax}
\providecommand{\bibinfo}[2]{#2}
\providecommand{\BIBentrySTDinterwordspacing}{\spaceskip=0pt\relax}
\providecommand{\BIBentryALTinterwordstretchfactor}{4}
\providecommand{\BIBentryALTinterwordspacing}{\spaceskip=\fontdimen2\font plus
\BIBentryALTinterwordstretchfactor\fontdimen3\font minus \fontdimen4\font\relax}
\providecommand{\BIBforeignlanguage}[2]{{%
\expandafter\ifx\csname l@#1\endcsname\relax
\typeout{** WARNING: IEEEtran.bst: No hyphenation pattern has been}%
\typeout{** loaded for the language `#1'. Using the pattern for}%
\typeout{** the default language instead.}%
\else
\language=\csname l@#1\endcsname
\fi
#2}}
\providecommand{\BIBdecl}{\relax}
\BIBdecl

\bibitem{endoscopyLim}
M.~Baumhauer, M.~Feuerstein, H.-P. Meinzer, and J.~Rassweiler, ``Navigation in endoscopic soft tissue surgery: Perspectives and limitations,'' \emph{Journal of Endourology}, vol.~22, no.~4, pp. 751--766, 2008.

\bibitem{endoscopyLim2}
U.~Kehler, J.~Regelsberger, and J.~Gliemroth, ``Pro and cons of different designs of rigid endoscopes,'' \emph{Minim Invasive Neurosurgery}, vol.~46, no.~4, pp. 205--207, 2003.

\bibitem{Mahboob}
M.~S. Eljamel and S.~O. Mahboob, ``The effectiveness and cost-effectiveness of intraoperative imaging in high-grade glioma resection; a comparative review of intraoperative ala, fluorescein, ultrasound and mri,'' \emph{Photodiagnosis Photodyn Ther}, vol.~16, pp. 35--43, 2016.

\bibitem{5ALA}
H.~A. Shah, S.~Leskinen, H.~Khilji, V.~Narayan, N.~Ben-Shalom, and R.~S. D'Amico, ``Utility of 5-ala for fluorescence-guided resection of brain metastases: a systematic review,'' \emph{J Neurooncol}, vol. 160, no.~3, pp. 669--675, 2022.

\bibitem{NF}
R.~Manoharan and J.~Parkinson, ``Sodium fluorescein in brain tumor surgery: Assessing relative fluorescence intensity at tumor margins,'' \emph{Asian J Neurosurg}, vol.~15, no.~1, pp. 88--93, 2020.

\bibitem{TIDTucker}
M.~Tucker, G.~Ma, W.~Ross, D.~M. Buckland, and P.~J. Codd, ``Creation of an automated fluorescence guided tumor ablation system,'' \emph{IEEE Journal of Translational Engineering in Health and Medicine}, vol.~9, pp. 1--9, 2021.

\bibitem{TZID}
T.~J. Zachem, J.~M. Komisarow, R.~A. Hachem, D.~W. Jang, W.~Ross, and P.~J. Codd, ``Intraoperative ex vivo pituitary adenoma subtype classification using noncontact laser fluorescence spectroscopy,'' \emph{J Neurol Surg B Skull Base}, vol.~84, no. S 01, p. P116, 2023.

\bibitem{NSGYlaser}
\BIBentryALTinterwordspacing
R.~W. Ryan, R.~F. Spetzler, and M.~C. Preul, ``Aura of technology and the cutting edge: a history of lasers in neurosurgery,'' \emph{Neurosurgical Focus FOC}, vol.~27, no.~3, p.~E6, 2009. [Online]. Available: \url{https://thejns.org/focus/view/journals/neurosurg-focus/27/3/article-pE6.xml}
\BIBentrySTDinterwordspacing

\bibitem{LITT}
K.~G. Holste and D.~A. Orringer, ``Laser interstitial thermal therapy,'' \emph{Neuro-Oncology Advances}, vol.~2, no.~1, 2019.

\bibitem{Biorender}
\BIBentryALTinterwordspacing
BioRender. (2023) Biorender. [Online]. Available: \url{BioRender.com}
\BIBentrySTDinterwordspacing

\bibitem{laserReview}
H.~C. Lee, N.~E. Pacheco, L.~Fichera, and S.~Russo, ``When the end effector is a laser: A review of robotics in laser surgery,'' \emph{Advanced Intelligent Systems}, vol.~4, no.~10, p. 2200130, 2022.

\bibitem{bendRadius}
A.~Salinas and J.~F.~T. Pittman, ``Bending and breaking fibers in sheared suspensions,'' \emph{Polymer Engineering \& Science}, vol.~21, no.~1, pp. 23--31, 1981.

\bibitem{hybridLaser}
D.~V.~A. Nguyen, C.~Girerd, Q.~Boyer, P.~Rougeot, O.~Lehmann, L.~Tavernier, J.~Szewczyk, and K.~Rabenorosoa, ``A hybrid concentric tube robot for cholesteatoma laser surgery,'' \emph{IEEE Robotics and Automation Letters}, vol.~7, no.~1, pp. 462--469, 2022.

\bibitem{ficheraLarynx}
A.~J. Chiluisa, N.~E. Pacheco, H.~S. Do, R.~M. Tougas, E.~V. Minch, R.~Mihaleva, Y.~Shen, Y.~Liu, T.~L. Carroll, and L.~Fichera, ``Light in the larynx: a miniaturized robotic optical fiber for in-office laser surgery of the vocal folds,'' in \emph{2022 IEEE/RSJ International Conference on Intelligent Robots and Systems (IROS)}, 2022, pp. 427--434.

\bibitem{hsmrLITT}
D.~Esser, J.~Peters, A.~Grillo, S.~Garrow, T.~Ball, R.~Naftel, D.~Englot, J.~Neimat, W.~Grissom, E.~Barth \emph{et~al.}, ``Robotic curvilinear laser thermal therapy probe for transforamenal hippocampotomy,'' in \emph{The Hamlyn Symposium on Medical Robotics}, 2022.

\bibitem{ficheraCHIP}
L.~Fichera, N.~P. Dillon, D.~Zhang, I.~S. Godage, M.~A. Siebold, B.~I. Hartley, J.~H. Noble, P.~T. Russell, R.~F. Labadie, and R.~J. Webster, ``Through the eustachian tube and beyond: A new miniature robotic endoscope to see into the middle ear,'' \emph{IEEE Robotics and Automation Letters}, vol.~2, no.~3, pp. 1488--1494, 2017.

\bibitem{bernardOCT}
J.~Yan, P.~Chen, J.~Chen, J.~Xue, C.~Xu, Y.~Qiu, H.~Fang, Y.~Lu, G.~K.~C. Wong, Y.-H. Liu, W.~Yuan, and S.~S. Cheng, ``Design and evaluation of a flexible sensorized robotic oct neuroendoscope,'' in \emph{2023 International Symposium on Medical Robotics (ISMR)}, 2023, pp. 1--7.

\bibitem{tumorCNC}
G.~Ma, W.~A. Ross, I.~Hill, N.~Narasimhan, and P.~J. Codd, ``A novel laser scalpel system for computer-assisted laser surgery,'' in \emph{2019 International Conference on Robotics and Automation (ICRA)}, 2019, pp. 386--392.

\bibitem{stereoCNC}
G.~Ma, W.~Ross, and P.~J. Codd, ``Stereocnc: A stereovision-guided robotic laser system,'' in \emph{2021 IEEE/RSJ International Conference on Intelligent Robots and Systems (IROS)}, 2021, pp. 540--547.

\bibitem{york}
P.~A. York, R.~Peña, D.~Kent, and R.~J. Wood, ``Microrobotic laser steering for minimally invasive surgery,'' \emph{Science Robotics}, vol.~6, no.~50, p. eabd5476, 2021.

\bibitem{tjGrasper}
T.~A. Brumfiel, R.~Qi, C.~Chapman, A.~Rashid, S.~N. Melkote, J.~J. Chern, and J.~P. Desai, ``Design and modeling of a sub-2 mm steerable neuroendoscopic grasping tool,'' \emph{IEEE Transactions on Medical Robotics and Bionics}, vol.~5, no.~4, pp. 1105--1109, 2023.

\bibitem{tendonE}
S.~Jeong, Y.~Chitalia, and J.~P. Desai, ``Design, modeling, and control of a coaxially aligned steerable (coast) guidewire robot,'' \emph{IEEE Robotics and Automation Letters}, vol.~5, no.~3, pp. 4947--4954, 2020.

\end{thebibliography}

\end{document}